\documentclass[12]{article}

\usepackage{arxiv}

\usepackage[utf8]{inputenc} 
\usepackage[T1]{fontenc}    
\usepackage{hyperref}       
\usepackage{url}            
\usepackage{booktabs}       
\usepackage{amsfonts}       
\usepackage{nicefrac}       
\usepackage{microtype}      
\usepackage{lipsum}

\usepackage{cite}
\usepackage{amsmath,amssymb,amsfonts}
\usepackage{algorithmic}
\usepackage{graphicx}
\usepackage{textcomp}
\usepackage{xcolor}
\usepackage{csquotes}
\graphicspath{{./figures/}}
\def\BibTeX{{\rm B\kern-.05em{\sc i\kern-.025em b}\kern-.08em
    T\kern-.1667em\lower.7ex\hbox{E}\kern-.125emX}}

\graphicspath{{./figures/}}
\usepackage{xcolor}

\title{Speech Driven Backchannel Generation using Deep Q-Network for Enhancing Engagement in Human-Robot Interaction}

\author{
  Nusrah Hussain, Engin Erzin, T. Metin Sezgin, and Y\"{u}cel Yemez \\
  College of Engineering, Ko\c{c} University\\
   Istanbul, Turkey\\
  \{ \textit{nhussain15, eerzin, mtsezgin,  yyemez}\}\textit{@ku.edu.tr} \\
}
\begin{document}

\maketitle

\begin{abstract}
We present a novel method for training a social robot to generate backchannels during human-robot interaction. We address the problem within an off-policy reinforcement learning framework, and show how a robot may learn to produce non-verbal backchannels like laughs, when trained to maximize the engagement and attention of the user. A major contribution of this work is the formulation of the problem as a Markov decision process (MDP) with states defined by the speech activity of the user and rewards generated by quantified engagement levels. The problem that we address falls into the class of applications where unlimited interaction with the environment is not possible (our environment being a human) because it may be time-consuming, costly, impracticable or even dangerous in case a bad policy is executed. Therefore, we introduce deep Q-network (DQN) in a batch reinforcement learning framework, where an optimal policy is learned from a batch data collected using a more controlled policy. We suggest the use of human-to-human dyadic interaction datasets as a batch of trajectories to train an agent for engaging interactions. Our experiments demonstrate the potential of our method to train a robot for engaging behaviors in an offline manner.
     
\end{abstract}


\section{Introduction}
Interaction of a robot with a human in a domestic environment needs to be characterized by behaviors and norms compatible with humans for a successful interaction experience. In applications like companionship, tutoring, and ambient assisting living, actions which encourage attention and engagement of the user are necessary for an effective interaction. A survey by Clavel et. al. summarizes the issues regarding engagement in human-agent interactions, emphasizing its importance and indicating the growing interest of researchers in the field \cite{clavel2016fostering}. Backchannels like non-verbal gestures (nods and smiles), non-verbal vocalizations (mm, uh-huh, laughs) and verbal expressions (yes, right) are an important aspect of engagement and have been shown to promote engagement and interest levels of the user \cite{turker2017analysis,inden2013timing}. Researchers have mainly focused on rule-based back-channel generation \cite{al2009generating,liu2012generation} or data-driven unsupervised methods \cite{admoni2014data}. In this work, we show how to formulate the problem in a reinforcement learning framework and train an agent to learn a policy for backchannel generation that maximizes the engagement of the user. To the best of our knowledge, this problem has not been addressed before in the context of reinforcement learning. 

Reinforcement learning (RL) has shown much success in the past few years as an optimization algorithm for problems having temporal structure and seems a natural choice for a variety of problems in robotics. Several works exist, which have used reinforcement learning to impart human-like behaviors into social robots. Qureshi et. al. \cite{qureshi2016robot,qureshi2017show} use multi-modal DQN to train a robot to greet like humans with the sequential actions of wait, look, wave and shake hand. The reward comes from successful handshakes via a sensor. RL is employed to adjust motion speed, timing, interaction distances, and gaze in the context of human-robot interaction (HRI) by Mitsunaga et. al. \cite{mitsunaga2006robot}. The reward is based on the amount of movement of the subject and the time spent gazing at the robot in one interaction. Lathuili{\`e}re et. al. \cite{lathuiliere2019neural} use recurrent neural network architecture in combination with Q-learning to find an optimal policy for robot gaze control in HRI. In these works, however, the agent either interacts with the environment (humans) for several days or training is done using simulators. In our work, we address the challenge where experience on a real physical system may be tedious to obtain, expensive, time-consuming and hard to simulate. We propose to use human-to-human interaction datasets as a batch of off-policy samples (trajectories) and use them in the context of offline batch reinforcement learning. The goal is to learn from the batch data the sequence of actions (or trajectories) that result in higher engagement and to formulate a policy around those regions of state-action space. Since we do not aim to mimic the dataset, instead of typical supervised learning metrics (like accuracy, recall, precision etc.) we evaluate our training using Bellman residual and expected return from the new policies.  
 
The contribution of this work is two fold: 
\begin{enumerate} 
	\item We propose a reinforcement learning formulation for backchannel generation that enhances engagement levels of the user during human-robot interaction. States are generated using speech features of the user and rewards are calculated by the engagement levels of the user. We demonstrate the experiments on laughs as a non-verbal backchannel. 
	\item We present the use of pre-recorded human-to-human dyadic interaction datasets as a batch of samples acquired by a behavior policy (another human). The optimal Q-value function is extracted from this batch data using a modified version of deep Q network (DQN) as a model-free value-based batch-RL algorithm. A greedy policy can implicitly be deduced from the optimal Q-value function.   
\end{enumerate}

\section{Related Work}
\subsection{Engagement in Interactions} \label{related_work_engagement}
Several definitions of engagement exist in the literature, which have been described in detail by Glas et. al. \cite{glas2015definitions}. Poggi describes engagement as: \enquote{the value that a participant in an interaction attributes to the goal of being together with the other participant(s) and of continuing the interaction} \cite{poggi2007mind}.  One of the pioneering studies on measurement of engagement is the work by Rich et. al. \cite{rich2010recognizing}, where the authors propose an engagement model for collaborative interactions between human and computer.  They define four types of events as engagement indicators, referred to as connection events (CEs), which include directed gaze, mutual facial gaze, adjacency pair, and backchannels. Directed gaze event is defined when both participants look at a nearby object related to the interaction at the same time. The mutual facial gaze occurs when there is face-to-face eye contact. Adjacency pair indicates a successful event when turn taking occurs with some minimal time gap. Finally, backchannels refer to the generation of audio-visual feedback by a listener during the speaker's turn. In our work, we use these connection events to quantify engagement and generate a single scalar value at each time step to represent the rewards. An alternative option may be to directly annotate the engagement levels in the dataset. However, automatic detection of engagement allows the refinement of the policy in the future by continuously updating the policy as the agent interacts with humans.

\subsection{Reinforcement Learning}
In general, the reinforcement learning formulates the optimization problem as a Markov decision process (MDP) $(S,A,p,r,\gamma)$ in which the environment has a state $s \in S$, the agent takes an action $a \in A$ and the scalar reward $r(s,a) \in \mathbb{R}$ is generated by the environment. The transition dynamics $p(s_{t+1}|s_t,a_t)$ gives the probability of next state $s_{t+1}$ given that at time $t$ the state $s_t$ is observed and action $a_t$ is taken. The discount factor $\gamma \in [0,1)$ weighs the future rewards, determining the extent of temporal data that is affected by the current action. The solution to any MDP is a policy $\pi(a|s)$ which maximizes the expectation of sum of discounted rewards, i.e., the return.  

Reinforcement learning is more commonly found as an online learning algorithm where the agent updates its policy while it interacts with the environment. However, RL may also be formulated as a batch (offline) or semi-batch learning technique, as described at length in a survey paper \cite{lange2012batch}. Batch reinforcement learning refers to the reinforcement learning setting where the task is to learn an optimal policy from a fixed batch of transitions sampled with some behavioral policy \cite{ernst2005tree}. Our interest in batch reinforcement learning is mainly due to the difficulty of human-robot interaction over extensive time lengths and under varying conditions. Moreover, several recordings of human-to-human dyadic interaction datasets are readily available. Thus, we utilize these recordings as a batch of samples collected by another policy and extract the optimum policy for our goal.

Neural fitted Q-iterations (NFQ) is a well-known batch reinforcement learning algorithm \cite{NFQ} based on Q-learning. It trains a neural network by fitting the Bellman optimality equation \cite{21sutton1998introduction} given by 
\begin{equation} \label{eq1}
Q^*(s,a) = \mathbb{E}[R_{t+1}+\gamma\underset{a'}{\max} Q^*(S_{t+1},a')|S_t=s,A_t=a].
\end{equation}

A more recent variant of the online Q-learning is the deep Q-network \cite{mnih2015human} which introduces two further concepts within the Q-learning approach. First, it uses experience replay to randomize over data in order to de-correlate sequential steps and secondly the target values, towards which the Q-values are iteratively updated, are refreshed periodically. Some works like \cite{hester2018deep} use DQN with demonstrations to accelerate the learning process. Similarly, we introduce the batch version of DQN (batch-DQN) and show its superiority over NFQ. We describe our batch-DQN technique in more detail in Section \ref{batch-DQN}.  

\section{Proposed Method}

\subsection{Dataset}
We work with the IEMOCAP dataset \cite{iemocap}, which is designed to analyze expressive human interactions. It consists of five sessions acted by ten professional actors performing dyadic human-to-human conversations. In total, there are 151 dialogs on 8 hypothetical scenes and 3 scripted plays, performed in pairs of the opposite gender. This dataset represents human behavior in a variety of situations where the policy behind each dialog may vary. 

We define our problem by assuming that, of the two actors, one represents the behavior policy (i.e., generates actions in the form of backchannels) while the second actor plays the role of the environment generating states and rewards. Considering the IEMOCAP dataset as a batch of trajectories collected by a behavioral policy, we apply our batch reinforcement learning method to extract an optimal policy that maximizes engagement during human-robot interaction.

\subsection{Markov Decision Process}
\label{MDP}
Though our framework is general for any type of event generation, we have conducted experiments to see the effectiveness of batch reinforcement learning on laughter generation. Given a state of the environment, the agent needs to make a decision on whether a backchannel event at that instance will contribute to engagement. We define the states, actions, and rewards as follows:
\begin{itemize}
	\item \textbf{State:} The state of the environment is represented by speech features extracted from past one second of data at every 25~msec step. This produces state information at a rate of 40~Hz. The dimension of the extracted feature is 209, which is described in detail in Section~\ref{Audio Features}. 
	\item \textbf{Action:} Agent's action is a binary variable, indicating the absence or presence of the backchannel. The backchannels of the user with behavioral policy in the dataset were labeled at a rate of 40~Hz.
	\item \textbf{Reward:} The reward is a scalar quantity which comes from the engagement measures of the user at every time step. The quantification of engagement from the dataset via Sidner's method is detailed in Section~\ref{engagement}.  
\end{itemize}  

\subsection{Generation of Tuples}
\begin{figure}[ht]
	\vskip 0.2in
	\begin{center}
		\centerline{\includegraphics[width=10cm,height=10cm,keepaspectratio]{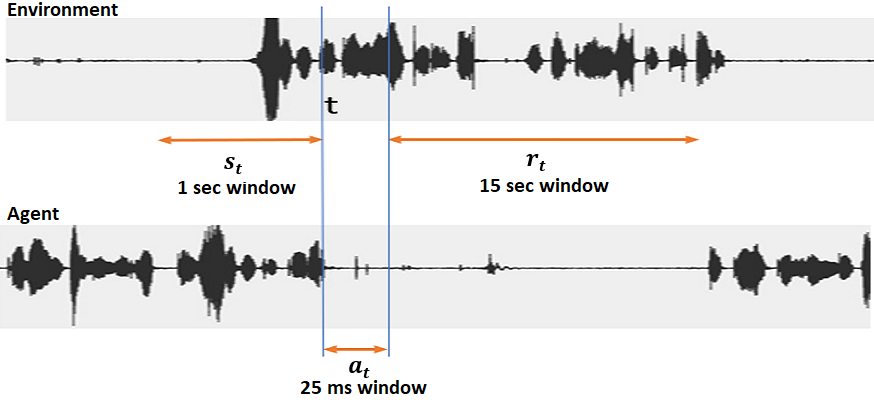}}
		\caption{Reinforcement learning formulation of speech driven backchannel generation (not drawn to scale)}
		\label{RLformulation}
	\end{center}
	\vskip -0.2in
\end{figure}

Batch reinforcement learning algorithms work with tuples of the form $\langle s_t,a_t,r_t,s_{t+1} \rangle$ for $t=1:T$. At time $t$, $s_t$ is the state of the environment, $a_t$ is the action of laughing by the agent and $r_t$ is the reward defined in terms of engagement levels generated from the environment. Figure \ref{RLformulation} shows the time windows used to extract states, rewards, and actions in one tuple. The dataset is pre-processed and such tuples are saved in a buffer.

\subsection{Speech Features}
\label{Audio Features}
The states are defined using the mel-frequency cepstrum coefficients (MFCCs) and prosody features extracted from the speech signal of the environment. 13-dimensional MFCC features are computed using 40 milliseconds sliding Hamming window at intervals of 25 milliseconds. The speech intensity, pitch, and confidence-to-pitch with their first derivates make up a 6-dimensional prosody feature, so a 19-dimensional feature vector is formed when MFCCs and prosody are concatenated as in \cite{Bozkurt2016}. Following this, feature summarization is performed where a set of statistical quantities are computed that describe the short-term distribution of each feature over the past one second. These quantities comprise eleven functions, more specifically mean, standard deviation, skewness, kurtosis, range, minimum, maximum, first quantile, third quantile, median quantile and inter-quartile range, which were successfully used before by  \cite{metallinou2013tracking}. The dimension of each of these statistical feature vectors is 11 times the dimension of the corresponding feature vector. This makes the feature size of length 209.

\subsection{Engagement Measurement}
\label{engagement}

Our measure of engagement is based on the method proposed in \cite{rich2010recognizing}, which is applicable to face-to-face collaborative HCI scenarios. Similar to the description in Section \ref{related_work_engagement}, we use the connection events (CE) (1) mutual facial gaze, (2) adjacency pair and (3) backchannels (that include laughs, smiles, nods and head-shakes) to quantify engagement. In \cite{rich2010recognizing}, the \enquote*{directed gaze} event is defined when the agent and the participant look at a nearby object related to the interaction at the same time. However, in our dataset, since we do not have objects of interest at which both parties look at, we exclude it in our definition. The extracted CEs are then used to calculate a summarizing engagement metric called \enquote*{mean time between connection events} (MTBCE). MTBCE measures the frequency of successful connection events that is for a given time interval T, MTBCE is calculated by T / (no. of CEs in T). As MTBCE is inversely proportional to engagement, similarly to \cite{rich2010recognizing}, we use pace = 1/MTBCE to quantify the engagement between a participant and the robot. The pace measure is calculated over a window of 15~seconds in our experiments.

\subsection{Batch-DQN}
\label{batch-DQN}
While the existing  batch RL techniques have presented their success in a number of settings, they do not efficiently scale to large datasets. When dealing with large amounts of patterns, the question arises whether all patterns need to be used in every training step and whether there exists a way in which only parts of the patterns can be selected while still allowing for successful training and good policies \cite{plutowski1993selecting}. Contrary to existing batch-RL techniques, we initialize an experience replay buffer much smaller than the batch data. During training, mini-batches of data are taken and only the samples that agree with current $\epsilon$-greedy policy are pushed into the buffer, where $\epsilon$ is the probability of choosing the action present in the batch and $(1-\epsilon)$ is the probability of choosing an action greedily. As $\epsilon$ decays with time, the batch converges to those samples that are more likely to be seen when following the optimal policy. Like DQN, at every iteration, a random mini-batch is sampled from the replay buffer and updates are performed using the Bellman control equation.

\begin{figure}[h!]
	\vspace{-0.1cm}
	\begin{center}
		\centerline{\includegraphics[width=10cm, height=10cm,keepaspectratio]{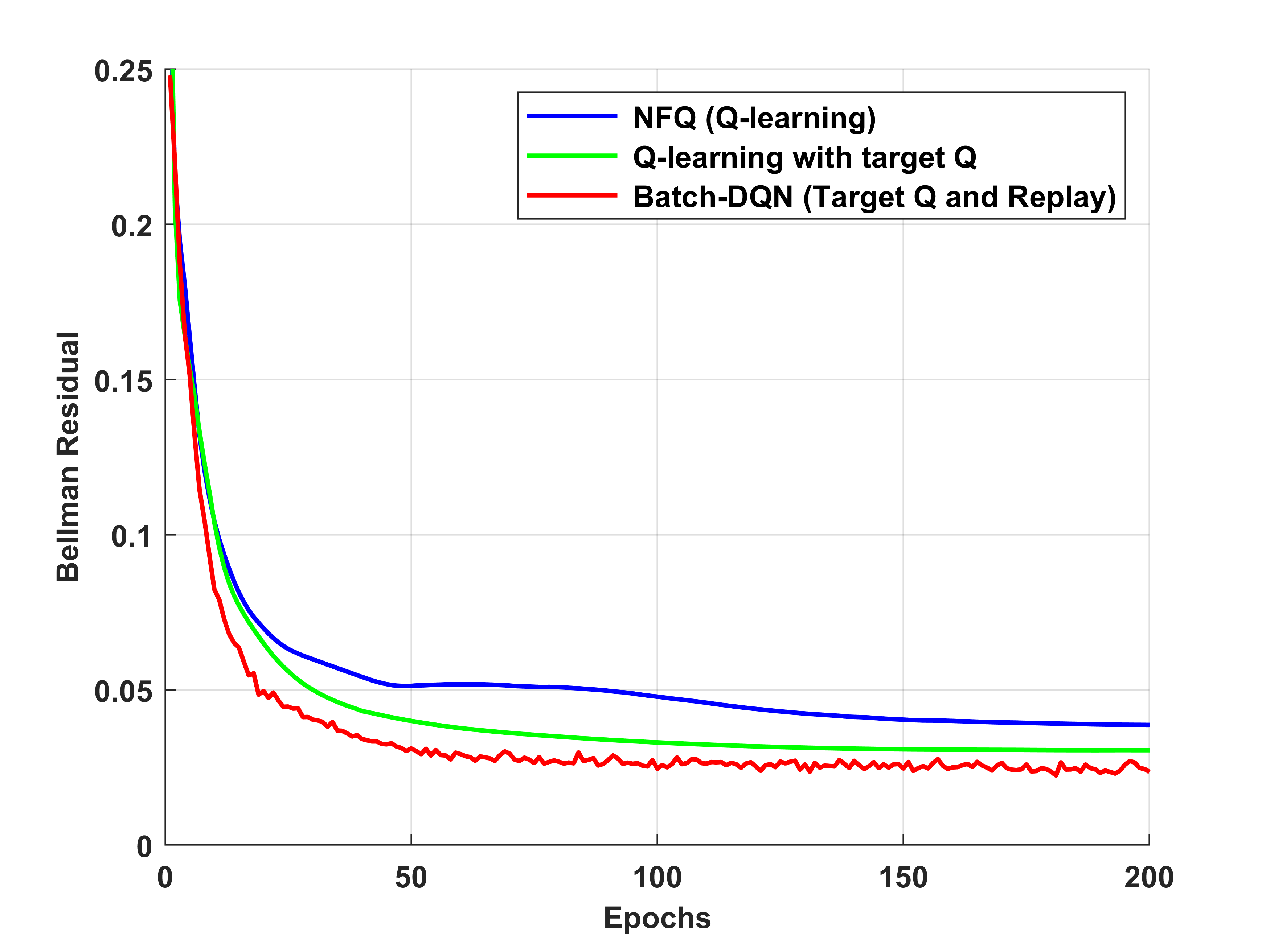}}
		\vspace{-0.1cm}
		\caption{Bellman residual vs training epochs}
		\label{bellman_residual}
	\end{center}
\end{figure}

\section{Experimental Setup}
From approximately 10 hours of recordings in IEMOCAP, $\langle s_t,a_t,r_t,s_{t+1} \rangle$ tuples form a batch data of size 1.5 million (as described in Section \ref{RLformulation}).  We further double the batch data by switching the roles of environment and behavior policy, hence increasing the batch size to approximately 3 million tuples. We define the train and test sets in the ratio 4:1 as leave one subject out (LOSO), hence 5 folds of training are performed which are subject independent. Training is done for the two techniques: batch-DQN and neural fitted Q-iterations (NFQ).  

We model the Q-function approximation network with a multi-layer perceptron of two hidden layers and use ReLU as the nonlinear activation function. The Q-network receives an input feature length 209, followed by hidden layers of sizes 100 and 25, and produces two Q-values so as to generate a backchannel or to remain inactive. The smooth-L1 loss from the Bellman control equation is minimized using Adam optimizer. We have used a discount factor value 0.99 in all experiments.

\subsection{Policy Evaluation Metrics}
Evaluation of the resultant policy is a challenging problem since the environment (i.e., the human participant) is not readily available in our case. Although it is possible to conduct experiments with human-robot interactions, it is desirable to first understand the policy's effectiveness using quantitative measures. Before presenting the results, we first describe below the metrics used for evaluation. 

\subsubsection{Bellman Residual}
For a Q-value function approximation network $Q_\theta$, the Bellman residual is defined as the difference between the two sides of a Bellman control equation \cite{baird1995residual}. A smaller residual error means that the learned policy is closer to the optimal policy and is a true Q-function since it follows the Bellman equation more closely. Similar to the work of \cite{ernst2005tree}, we compute the Bellman residual, $B_r$, over the entire batch of data as
\begin{equation}
\label{bellmaneq}
B_r=\frac{1}{|\mathcal{B}|}\sum_{ \mathcal{B}} ( Q_{\theta}(s_t,a_t) - [ r_t + \gamma*\underset{a \in A}{\max} Q_{\theta}(s_{t+1},a) ])^2 .
\end{equation} 

\subsubsection{Off-Policy Policy Evaluation (OPE)} 
Off-policy policy evaluation (OPE) is defined as the problem of estimating the value $V^\pi$ of a new policy using the batch of samples collected independently from a behavior policy. In the last few years, many OPE techniques have emerged because of its importance in cases where a new policy cannot be tested directly with the environment \cite{thomas2016data,doroudi2017importance}. To compare the values of policies resulting from NFQ and batch-DQN, we applied the step-wise weighted importance sampling estimator (WIS) given by
\begin{equation}
\label{eq:ope}
\hat{V}^{\pi}_{step-WIS}=\sum_{i=1}^n\sum_{t=0}^{T-1}\gamma^t\frac{\rho_t^{(i)}}{\sum_{i=1}^n \rho_t^{(i)}}r_t^{(i)},
\end{equation}
where $n$ is the number of trajectories, $T$ is the length of each trajectory and $\gamma$ is the discount factor. Then, the importance weight $\rho$ is defined as the ratio of the probability of the first $t + 1$ steps of a trajectory under $\pi$ to the probability under a behavior policy $\pi_b$ and is given as 
$\rho_t = \prod_{i=0}^{t} \frac{\pi(a_i|s_i	)}{\pi_b(a_i|s_i)}$. The importance sampling approach to evaluation relies on using the importance weights $\rho_t$ to adjust for the difference between the probability of a trajectory under the behaviour policy $\pi_b$
and the probability under the evaluation policy $\pi$. To perform this evaluation we defined trajectories as frames of length $T=100$ with shifts of one sample. Following discussion of the work in \cite{raghu2018behaviour}, the behavior policy $\pi_b$ was estimated using approximate nearest neighbor \cite{Hyvonen2016}.

\subsubsection{Naturalness of laughs}
Another metric we take into consideration is similarity of laughs to that of a human. For example, a policy that results in laughs which lasts several minutes would not be natural and would result in discomfort of the user. In order to assess the naturalness of the laughs generated for the agent, we analyze the distribution of laughter duration present in the dataset. A successful agent should be able to produce similar statistics. To measure the similarity between two probability histograms we used symmetric Kullback-Leibler Divergence measure as well as statistical metrics like mean, max and inter-quartile range.
  
\section{Results \& Discussion}	
To illustrate the effectiveness of batch-DQN over NFQ, Figure~\ref{bellman_residual} shows the Bellman residuals plotted against the epoch number for one fold of the laughter training. We observe that the error reduces when a separate target network is introduced, and improves further when samples are randomly selected from the replay buffer which contains tuples complying with the current epsilon-greedy policy. A closer look at the minimum Bellman residuals when averaged over all folds of the training shows the superiority of batch-DQN. Table~\ref{losses} gives the mean errors obtained over the test sets.

\begin{table}[h!]
	\caption{Bellman residuals for test set over 5-fold training}
	\label{losses}
	\vspace{-0.2cm}
	\begin{center}

			\begin{tabular}{lll}
				\toprule
				\textbf{Technique} & \textbf{Bellman Residual} \\
				\midrule					
				Neural Fitted Q-Iterations & 	$0.0571 \pm 0.0097$ \\
				Batch-DQN  &$ 0.0371 \pm 0.0130 $ \\
				\bottomrule
			\end{tabular}

	\end{center}
	\vskip -0.1in
\end{table} 
\vspace{-0.1cm}

The off-policy policy evaluation results obtained from step-wise weighted importance sampling are shown in Table~\ref{ope}. The behavior policy is modelled with approximate nearest neighbor algorithm while the policies learned from NFQ and batch-DQN are deduced from the corresponding Q-networks such that 0.9 probability is assigned to the action (laughter or non-laughter) suggested by the greedy policy. The values returned by OPE represent the expected cumulative discounted rewards estimated for each method. Ideally these values need to be computed over infinite lengths but keeping in mind the numerical limitations we calculated them over trajectories of length 250 samples. While both techniques perform better than the behavior policy baseline, it can be noted that the policy obtained by our batch-DQN method outperforms the policy learned by the NFQ technique.

\begin{table}[h!]
	\caption{Off-policy policy evaluation (OPE) results}
	\label{ope}
	\vspace{-0.2cm}
	\begin{center}
		\begin{small}
			\begin{tabular}{lll}
				\toprule
				\textbf{Technique} & \textbf{Estimated $V^\pi$} \\
				\midrule
				Dataset (Behavior Policy) & 23.15 \\
				Neural Fitted Q-Iterations & 24.32\\
				Batch-DQN  & $\textbf{27.57}$ \\
				\bottomrule
			\end{tabular}
		\end{small}
	\end{center}
\end{table}

Finally, Table~\ref{laugh_table} shows the statistical similarity of laughter duration generated by each policy to that of a human. While the mean is comparable to a human, the prominent weakness of NFQ policy is seen by the maximum length of the laugh it generates.  NFQ results in laughs many folds longer than a typical human laugh ($\sim$ 198 sec). A closer look at the reward distribution in the dataset reveals that the mean reward is higher in the intervals where laughs are present. This explains the reason why both techniques prefer to generate laughs more frequently than in the dataset. This can be improved upon in the future work by imposing some form of constraints to make the lengths of laughs more natural. The KL-divergence value also shows the superiority of batch-DQN over NFQ.  
 
\begin{table}[!h]

	\caption{Similarity of agent laughs to human laughs}
	\label{laugh_table}	
	\begin{center}
	\vspace{-0.2cm}
		\begin{tabular}{p{2.5cm}p{1cm}p{1cm}p{1cm}p{3cm}}
			\toprule
			& \multicolumn{3}{p{3.7cm}}{\textbf{Laugh Duration (sec)}}  &  \textbf{KL-Divergence} \\
			& Mean & Max & Std & \\
			\midrule					
			Human & 0.95 & 9.17 & 0.89 &  - \\
			NFQ & 1.15 & 198.45 & 3.18 & 0.2890	  \\
			Batch-DQN  & 0.76 & 46.45 & 0.99 & 0.1921 \\
			\bottomrule
		\end{tabular}
		
	\end{center}
\end{table}

\section{Conclusion}
We have presented a scheme to train a robot for backchannel generation in a human-robot interaction, so as to maximize the engagement of the user. The formulation of this problem as a reinforcement learning problem allows the robot to learn the effect of an action in current time step on the entire future interaction. We have also shown how the available datasets on human-to-human interaction may be used as a batch of off-policy trajectories. Our experiments have shown the advantage of our batch-DQN algorithm over neural fitted Q-iterations technique which can be considered as a baseline for batch RL. An immediate extension of this work is to perform subjective evaluations with human participants. Furthermore, this work may be extended by enriching the state definition using visual features and by training for other forms of backchannels.   

\bibliographystyle{IEEEtran}

\end{document}